\documentclass{article}

\PassOptionsToPackage{sort, numbers, compress}{natbib}

\usepackage[preprint]{neurips_2024}

\usepackage[utf8]{inputenc} 
\usepackage[T1]{fontenc}    
\usepackage{url}            
\usepackage{booktabs}       
\usepackage{amsfonts}       
\usepackage{nicefrac}       
\usepackage{microtype}      
\usepackage[table]{xcolor}  
\usepackage{xcolor}         
\usepackage{graphicx}
\usepackage{multirow}
\usepackage{tabularx}
\usepackage{pifont}
\usepackage{rotating}
\usepackage{arydshln}
\usepackage{makecell}
\usepackage{nicematrix,tikz}
\usepackage[colorlinks=true, urlcolor=blue]{hyperref}

\newcommand{\cmark}{\ding{51}}%

\definecolor{lightgray}{gray}{0.9}
\definecolor{LightCyan}{rgb}{0.88,1,1}
\newcolumntype{a}{>{\columncolor{lightgray}}c}
\newcolumntype{b}{>{\columncolor{white}}c}

\title{OpenTAD: A Unified Framework and Comprehensive Study of Temporal Action Detection}

\author{
Shuming Liu\thanks{Co-First Authors.}  \quad
Chen Zhao\footnotemark[1]  \quad
Fatimah Zohra \quad
Mattia Soldan \quad
Alejandro Pardo \\
\textbf{Mengmeng Xu} \quad
\textbf{Lama Alssum} \quad
\textbf{Merey Ramazanova} \quad
\textbf{Juan León Alcázar} \\
\textbf{Anthony Cioppa} \quad
\textbf{Silvio Giancola} \quad 
\textbf{Carlos Hinojosa} \quad
\textbf{Bernard Ghanem} \\\\
Video Understanding Group, Image and Video Understanding Lab (IVUL)\\ 
King Abdullah University of Science and Technology (KAUST)}

\begin{document}
\maketitle
\begin{abstract}
Temporal action detection (TAD) is a fundamental video understanding task that aims to identify human actions and localize their temporal boundaries in videos.
Although this field has achieved remarkable progress in recent years, further progress and real-world applications are impeded by the absence of a standardized framework.
Currently, different methods are compared under different implementation settings, evaluation protocols, etc., making it difficult to assess the real effectiveness of a specific technique.
To address this issue, we propose \textbf{OpenTAD}, a unified TAD framework consolidating 16 different TAD methods and 9 standard datasets into a modular codebase. In OpenTAD, minimal effort is required to replace one module with a different design, train a feature-based TAD model in end-to-end mode, or switch between the two.
OpenTAD also facilitates straightforward benchmarking across various datasets and enables fair and in-depth comparisons among different methods.
With OpenTAD, we comprehensively study how innovations in different network components affect detection performance and identify the most effective design choices through extensive experiments.
This study has led to a new state-of-the-art TAD method built upon existing techniques for each component. We have made our code and models available at \url{https://github.com/sming256/OpenTAD}. 
\end{abstract}
\section{Introduction}
\label{sec:intro}
With the rapid growth of video content across diverse domains—ranging from social media and entertainment to surveillance and autonomous systems—understanding and analyzing videos has become a crucial research focus.  
While significant progress has been made in classification tasks such as object and action recognition, temporal reasoning remains a challenging problem~\cite{wu2019long, wu2021towards, lin2022egocentric, zhang2022actionformer, zhao2023re2tal, ren2023timechat}. In particular, \textbf{temporal action detection} (TAD)~\cite{zhao2023re2tal,zhang2022actionformer,lin2019bmn,xu2020g} has emerged as a fundamental task, serving as a pretext for various downstream applications, including dense video captioning and natural language temporal video grounding~\cite{soldan2021mad,Hendricks2017LocalizingMI,grauman2022ego4d, chen2022internvideo}. 

Temporal action detection aims to locate the start and end timestamps of each action instance as well as to identify their categories within a video sequence~\cite{zhao2023re2tal,zhang2022actionformer,lin2019bmn,xu2020g}.
Formally, given a video sequence of $T$ frames $\mathcal{V}=\{I_t \in \mathbb{R}^{H \times W \times 3}\}_{t=1}^T$, TAD predicts a set $M$ of action segments: $\left \{ \left ({t}_{m, s},{t}_{m, e},  {s}_m \right ) \right \}_{m=1}^{M}$, where ${t}_{m, s}$ and ${t}_{m, e}$ denote the start and end timestamps of an action, respectively, and ${c}_m$ represents the action label. 
Conceptually, TAD is analogous to the object detection task~\cite{girshick2015fast,ren2016faster,carion2020end,redmon2016you,zhang2023dino}, which regresses object boundaries as 2D bounding boxes, whereas TAD regresses action boundaries along the temporal dimension of a video. 

Various deep neural network-based methods have been proposed for TAD~\cite{wang2023temporal,hu2024overview,vahdani2022deep,escorcia2016daps,xu2017r,chao2018rethinking,zhang2022actionformer,liu2022end,zeng2019graph,xu2020g}, continuously advancing detection performance. 
Early approaches, such as DAPS~\cite{escorcia2016daps} and R-C3D~\cite{xu2017r}, pioneered proposal-based methods by adapting techniques from the image domain, particularly the two-stage object detection framework derived from Faster R-CNN~\cite{ren2015faster}. Subsequent works introduced techniques to improve the quality of proposals, focusing on refining boundary localization and exploring alternative loss functions~\cite{lin2019fast}. 
Meanwhile, other methods have tackled challenges unique to the video domain, such as the large variation in action durations~\cite{chao2018rethinking}. These variations make it difficult to apply fixed-scale receptive fields, prompting researchers to develop multi-scale modeling techniques that dynamically adapt to different temporal extents.

More recently, recognizing the imbalanced development between video temporal action detection and image object detection techniques, researchers have sought to adapt innovations from object detection to TAD. This has led to the emergence of one-stage methods~\cite{zhang2022actionformer, shi2023tridet, tang2023temporalmaxer, liu2024harnessing, yang2024dyfadet, lin2021learning, ning2021srf, kang2022htnet} and DETR-based architectures~\cite{liu2022end, zhu2024dual, kim2023self, kim2024prediction}, which leverage direct action localization without requiring predefined proposals.
Another line of research has incorporated emerging network architectures, including graph convolutional networks (e.g., GTAD~\cite{xu2020g}, P-GCN~\cite{zeng2019graph}, VSGN~\cite{zhao2021video}), Transformers (e.g., ActionFormer~\cite{zhang2022actionformer}, ViT-TAD~\cite{yang2024adapting}, LIP~\cite{kim2024long}), and Mamba models (e.g., VideoMambaSuite~\cite{chen2024video}, S-Temba~\cite{sinha2025ms}). These architectures have significantly improved the ability to model long-range temporal dependencies and capture fine-grained action boundaries.
Furthermore, recent studies suggest that end-to-end training generally yields higher accuracy than feature-based training. To address the memory constraints when using large-scale video encoding networks, researchers have introduced various optimization strategies, as seen in E2E-TAD~\cite{liu2022empirical}, Re$^2$TAL~\cite{zhao2023re2tal}, and AdaTAD~\cite{liu2024adatad}.

Throughout the brief history of TAD, different methods have focused on distinct aspects of the TAD pipeline, including video representation backbones, detection heads, loss functions, and training strategies. Consequently, a fair comparison between these methods requires evaluation under a unified TAD framework.
Unfortunately, existing methods are often implemented in different frameworks, with varying experimental setups and evaluation protocols. As a result, it becomes difficult to determine whether a reported performance gain originates from the proposed innovations, a more effective implementation framework, or simply hyperparameter tuning. 
To further advance this field, a unified TAD framework is needed—one that holistically evaluates each individual design choice while ensuring fair and reproducible comparisons across different approaches.

This paper introduces \textbf{OpenTAD}, an \textbf{Open}-sourced unified framework for \textbf{T}emporal \textbf{A}ction \textbf{D}etection, accompanied by a comprehensive study of various innovations across different TAD components through extensive experiments. OpenTAD integrates a diverse suite of TAD methods and datasets within a single framework and codebase, facilitating streamlined implementation, fair comparisons, rigorous analysis, and efficient benchmarking.
By leveraging OpenTAD, researchers can systematically identify key factors influencing performance—whether stemming from specific module designs, training strategies, or data processing techniques—thereby enabling the development of more advanced TAD models by combining the strengths of existing approaches. Additionally, OpenTAD provides a scalable foundation for extending existing methods to a broader range of datasets and application scenarios. 
We believe that OpenTAD will foster faithful and efficient development, evaluation, and assessment of novel module designs for TAD. Our contributions are as follows.

\begin{figure}[t]
\begin{center}
\footnotesize
\includegraphics[width=0.99\textwidth]{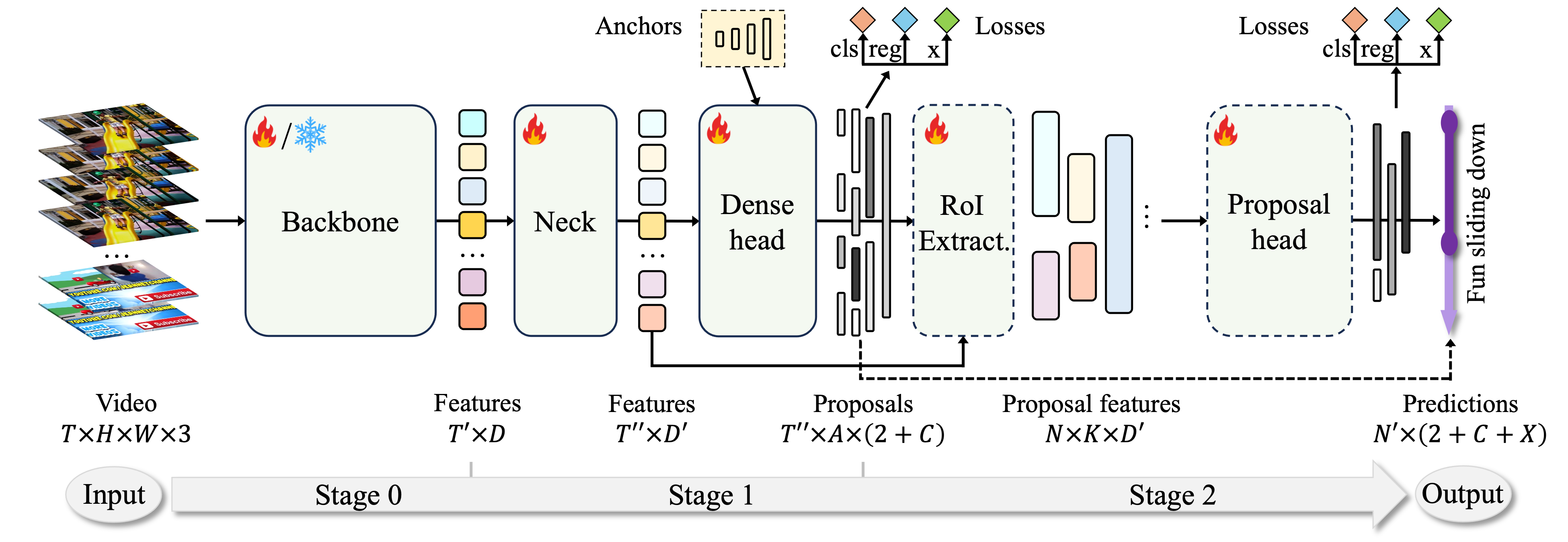}
\end{center}
\caption{\textbf{Unified TAD Pipeline.} Recent TAD methods follow this three-step framework to predict action classes and start/end timestamps from input videos. \textbf{1) Stage 0:} Videos are encoded into features using a pretrained video backbone, which may be either fine-tuned or frozen during training. \textbf{2) Stage 1:} This stage consists of a neck for temporal aggregation of snippet-level features and a dense head that generates snippet-level predictions. \textbf{3) Stage 2 (optional):} This stage further refines action segment proposals using RoI extraction and produces per-proposal predictions.}
\vspace{-5pt}
\label{fig:tad_arch}
\end{figure}

\begin{itemize}
    \item[(I)] We introduce \textbf{OpenTAD}, a unified framework that modularizes the process of temporal action detection. OpenTAD standardizes the implementation of diverse TAD methods, datasets, and evaluation metrics, enabling systematic assessment, in-depth analysis, and extensive benchmarking. With OpenTAD, researchers can seamlessly switch between feature-based and end-to-end training methods based on computational budgets.
    
    \item[(II)] Based on the OpenTAD framework, we provide an open-source code suite, where we have re-implemented \textbf{16 diverse methods}, including one-stage, two-stage, DETR-based, and end-to-end approaches, as well as \textbf{9 benchmark datasets}. Thanks to its unified design, OpenTAD can be easily extended to support additional methods and datasets, and we remain committed to continuously expanding its capabilities. 

    \item[(III)] We conduct extensive experiments within OpenTAD to comprehensively analyze various architectural and methodological innovations across different TAD components and datasets. Our study identifies the most effective module designs, and by systematically integrating these into existing TAD methods, we achieve new state-of-the-art performance.
\end{itemize}

\section{A Unified TAD Framework}
We unify modern temporal action detection methods within a single framework by analyzing the essential roles of different design choices and modularizing the entire pipeline, as illustrated in Fig.~\ref{fig:tad_arch}. 
To predict action categories and start/end timestamps from an input video, the TAD framework consists of three stages of network components, which will be detailed in Sections~\ref{sec:stage0}, \ref{sec:stage1}, and \ref{sec:stage2}. Following the introduction of OpenTAD’s core components, we present our unified data preprocessing and postprocessing pipeline in Sec.~\ref{sec:data_processing}. Finally, we describe how various TAD methods are integrated into OpenTAD, categorizing them into one-stage, two-stage, DETR-based, and end-to-end approaches in Sec.~\ref{sec:categorization}. 

OpenTAD is implemented in PyTorch within a unified codebase, which we have open-sourced along with our trained models, datasets, and configurations for all re-implemented methods. In the OpenTAD codebase, each component is designed as an independent module, allowing it to be instantiated in different variants without affecting—or being affected by—other components. This modular design enables seamless integration of novel techniques into the framework while ensuring fair comparisons under a consistent setup.

For clarity, we consistently adopt the following notations to represent the dimensions of video data throughout the OpenTAD pipeline. We use $H$ and $W$ for frame height and width, respectively, $T$ for sequence length, $D$ for the number of feature channels, $C$ for the number of action classes, $A$ for the number of predefined candidate actions per temporal location (referred to as \textit{anchors}), $N$ for the total number of predicted candidate actions (including \textit{proposals} and final predictions), and $K$ for the temporal length of each extracted proposal. The prime notation ($'$) is used to indicate a value change in a specific dimension.

\subsection{Stage 0: Video Feature Extraction}\label{sec:stage0}

The first computational block of the TAD pipeline encodes raw videos into snippet-level or frame-level features, which serve as generic video representations requiring further processing by subsequent TAD-specific components. Therefore, we designate this step as \textbf{Stage 0}. 

\textbf{Backbone: $T\times H\times W\times3\rightarrow T'\times D$}. The backbone is responsible for feature extraction, encoding an input video sequence $\mathcal{V}$ into a sequence of feature vectors $\mathcal{F}$. It consists of multiple layers that perform spatio-temporal aggregation, such as 3D convolution (e.g., I3D~\cite{carreira2017quo}, SlowFast~\cite{slowfast}) or space-time attention (e.g., VideoSwin~\cite{vswin}, VideoMAE~\cite{tong2022videomae}). This backbone is typically pretrained on a large-scale dataset of short video clips, such as Kinetics~\cite{zisserman2017kinetics}, using classification tasks, self-supervised learning~\cite{tong2022videomae}, or vision-language pretraining~\cite{li2023unmasked}. 

We identify two primary approaches for encoding video sequences using the backbone: (1) \textbf{snippet encoding.} A snippet refers to a short video clip consisting of a small number of frames. In this approach, the video sequence is divided into $T'$ snippets, typically with overlap, and each snippet is processed independently by the backbone to generate a feature vector, i.e., temporal aggregation occurs only within each snippet. Both the spatial and temporal dimensions are globally pooled in the output feature vector for each snippet. (2) \textbf{frame encoding}. In contrast, this approach processes the entire video sequence $\mathcal{V}$ as a single long clip, meaning that all frames are temporally-aware within the backbone. The spatial dimension is globally pooled, while the temporal dimension $T$ is preserved in the output feature vectors. Frame encoding offers greater computational and memory efficiency by reducing redundant computations across neighboring snippets, making it particularly suitable for end-to-end training (Sec.~\ref{sec:categorization}). On the other hand, snippet encoding processes each snippet independently, making it more appropriate for offline feature extraction in feature-based TAD settings. OpenTAD supports both approaches, ensuring flexibility for different experimental setups.

\subsection{Stage 1: Temporal Aggregation and Initial Prediction}\label{sec:stage1}

This stage consists of a neck component for temporal aggregation of video features and a dense head that directly predicts candidate actions. It corresponds to the first stage in two-stage methods (e.g., BMN~\cite{lin2019bmn}, GTAD~\cite{xu2020g}) and the sole stage in one-stage methods (e.g., ActionFormer~\cite{zhang2022actionformer}, TriDet~\cite{shi2023tridet}). Therefore, we designate this step as \textbf{Stage 1}.

\textbf{Neck: $T'\times D\rightarrow T''\times D'$.} The neck performs temporal aggregation across the entire feature sequence $\mathcal{F}$. Compared to the backbone, it is more lightweight and can provide either single-scale~\cite{xu2020g, lin2019bmn} or multi-scale~\cite{zhao2021video, zhang2022actionformer} temporal features. The neck is a crucial component for TAD as it consists of temporal aggregation modules, such as 1D convolution~\cite{lin2018bsn, lin2019bmn}, graph networks~\cite{xu2020g}, temporal attention~\cite{zhang2022actionformer}, or state-space models (SSM)~\cite{chen2024video}, to capture complex temporal relations.
In Sec.~\ref{sec:component_importance}, we analyze the importance of the neck and compare the effectiveness of different design choices.

\textbf{Anchor Generation:} Anchors~\cite{redmon2016you} are predefined candidate action segments at each temporal location, allowing the model to predict offsets relative to these anchors instead of directly estimating action boundaries. This strategy simplifies training and improves localization accuracy. Anchors can be defined in different formats, such as fixed temporal segment sizes centered at each location~\cite{zhao2021video} or all possible start-end combinations~\cite{xu2020g}. 
Recent one-stage methods have shifted towards an anchor-free approach, where the action start and end offset are directly predicted~\cite{zhao2021video,zhang2022actionformer}.

\textbf{Dense Head: $T''\times D' \rightarrow T''\times \left(2A+CA+X\right)$.} The dense head is responsible for making predictions at each temporal location after the neck. In one-stage methods, these predictions are directly used for post-processing, whereas in two-stage methods, they serve as \textit{proposals} to be further refined in the next stage. We categorize the predicted outputs at each temporal location as follows: (1)~\textit{start/end offsets:} $2A$, where $A$ represents the number of anchors at each temporal location. These offsets define the distances of ground truth actions relative to predefined anchors. (2)~\textit{confidence scores of each category:} $CA$, where $C$ denotes the number of action categories. In some cases, $C$ is set to 2 to represent the presence or absence of an action. (3) \textit{other auxiliary predictions,} which include probabilities of being inside an action segment or located on a start/end boundary~\cite{lin2018bsn, xu2020g}. These prediction branches are optimized using different loss functions, \textit{e.g.}, GIoU loss~\cite{rezatofighi2019generalized} for the start/end offsets regression~\cite{zhao2021video} and cross-entropy loss or focal loss~\cite{Lin2020FocalLF} for category prediction~\cite{zhang2022actionformer}.

\subsection{Stage 2: RoI Extraction and Action Refinement}\label{sec:stage2}

Given the predicted action candidates from stage 1, i.e., proposals, this stage further refines their boundaries and confidence scores to yield higher-quality action predictions. To achieve this, two key components are required: RoI extraction and the proposal head. This stage is unique to two-stage methods~\cite{lin2019bmn,xu2020g} and is therefore designated as \textbf{Stage 2}.

\textbf{RoI Extraction: $T''\times D' \rightarrow N\times K \times D'$.} Given the predicted action candidates, the corresponding proposal features are constructed through the region of interest (RoI) extraction module. Specifically, features within the proposal boundaries are extracted to generate $N$ proposal features, each with a temporal length of $K$ and a feature dimension of $D'$.
Different RoI extraction methods process features differently. For example, RoI Align~\cite{chao2018rethinking} and SGAlign~\cite{xu2020g} utilize all features within the boundaries, whereas PBRNet~\cite{liu2020progressive} and VSGN~\cite{zhao2021video} extract only center and boundary features. Additionally, the boundary matching mechanism in BMN~\cite{lin2019bmn} can also be regarded as a matrix implementation of RoI alignment.

\textbf{Proposal Head: $N\times K \times D' \rightarrow N\times (2 + C + X')$.} The proposal head processes each proposal feature to refine its boundaries and predict the action category~\cite{zhao2021video, lin2021learning}. Some methods also include additional predictions, such as completeness scores~\cite{lin2019bmn,xu2020g}. To further improve localization accuracy, multiple proposal heads can be stacked sequentially to form a cascade refinement~\cite{liu2020tsi}.
When dealing with a large number of proposals, the computational complexity of the proposal head can significantly exceed that of the dense head. This is one of the primary reasons why one-stage methods are preferred in scenarios where efficiency is a priority. Similar to the dense head, regression loss and classification loss are used to supervise boundary refinement and action classification, respectively, although different label assignment strategies are often employed.

\subsection{Data Pre-processing and Post-processing}\label{sec:data_processing}

Beyond the neural network components, pre-processing of video data and post-processing of proposal predictions play crucial roles in improving TAD performance.

\textbf{Pre-processing:} Since the TAD task involves long video sequences with variable durations, each input video sequence must be mapped to a fixed length $T$ for batch processing during training. In OpenTAD, we support three temporal-scale mapping mechanisms to accommodate various methods: (1) Rescaling the entire video sequence to a fixed length of $T$ frames/features, e.g., via interpolation. This approach is often used in datasets such as ActivityNet~\cite{caba2015activitynet} and HACS~\cite{zhao2019hacs}, where action durations vary significantly and may span almost the entire video. (2) Randomly extracting a continuous video segment of $T$ frames from a long video. In this setting, a subset of frames is used for training in each epoch, as seen in ActionFormer on THUMOS dataset~\cite{zhang2022actionformer}. If the video sequence is shorter than $T$, zero-padding is applied to reach the required length. (3) Applying a sliding window to partition each video into multiple segments, resulting in multiple training samples from a single video per epoch~\cite{lin2019bmn,xu2020g}.

\textbf{Post-processing:} Once action predictions are generated from either the dense head or the proposal head—typically numbering in the thousands—redundant proposals must be reduced. To achieve this, techniques such as non-maximum suppression (NMS) or Soft-NMS~\cite{softNMS} are commonly employed. If a video has been partitioned into multiple segments using a sliding window during pre-processing, all predictions from these segments are first aggregated before applying NMS or Soft-NMS.

\subsection{TAD Methods and Datasets Unified in OpenTAD }\label{sec:categorization}

As a unified framework, OpenTAD supports a wide range of deep learning-based TAD methods. We have modularized and re-implemented \textbf{16 TAD methods} under the OpenTAD framework and provide a detailed mapping of how each method instantiates OpenTAD components in Tab.~\ref{tab:mapping_to_opentad} in the supplementary material. Additionally, we compare our re-implementation performance to the original reported results for all methods in Tab.~\ref{tab:re-implementation_results}, and present benchmarking results on two datasets in Tab.~\ref{tab:benchmark_anet} and Tab.~\ref{tab:benchmark_thumos}, also in the supplementary material.

These methods belong to various categories, each contributing to different components, as detailed in Table~\ref{tab:supported_methods}. We categorize a TAD method as \textbf{end-to-end} if it jointly trains the backbone alongside the other network components, \textit{e.g.}, AFSD~\cite{lin2021learning}, Re$^2$TAL~\cite{zhao2023re2tal}, and AdaTAD~\cite{liu2024adatad}. Conversely, we classify a method as \textit{feature-based} if it utilizes a pretrained and frozen backbone during training, \textit{e.g.}, BMN~\cite{lin2019bmn} and ActionFormer~\cite{zhang2022actionformer}. Feature-based methods are further divided into \textit{one-stage} or \textit{two-stage}, depending on whether they incorporate Stage 2 components. Additionally, we classify methods as \textit{DETR-based} if they adopt the encoder-decoder architecture of DETR to directly learn action queries (e.g., TadTR~\cite{liu2022end}).
In Column 4 of Table~\ref{tab:supported_methods}, we indicate the OpenTAD component corresponding to each paper’s primary contribution, while in Column 5, we describe the claimed innovations related to that component.

\begin{table}[t]
\centering
\caption{\textbf{Example TAD methods unified in OpenTAD with their main contribution.} \textbf{Column 4}: the OpenTAD component
corresponding to each paper’s primary contribution.  \textbf{Column 5}: the claimed innovations related to the target component.}
\small
\setlength{\tabcolsep}{6.5pt}
\begin{Tabular}{@{}lllll@{}}
\toprule
\textbf{Category}    & \textbf{Method}           & \textbf{Publication} & \textbf{Target} & \textbf{Innovation}   \\
\midrule
\multicolumn{5}{c}{\textbf{Feature-based Approaches}} \\
\addlinespace[2pt]
\hdashline
\addlinespace[4pt]
\multirow{6}{*}{\makecell[l]{ One-stage}}   & ActionFormer~\cite{zhang2022actionformer}     & ECCV 22   & Neck          &   Transformer                 \\
            & TriDet~\cite{shi2023tridet}           & CVPR 23   & Neck; Head          &   SGP; Trident Head                 \\
            & TemporalMaxer~\cite{tang2023temporalmaxer}    & ArXiv 23    & Neck         &     MaxPooling               \\
            & VideoMambaSuite~\cite{chen2024video}  & ArXiv 24    & Neck          &     Mamba               \\
            & DyFADet~\cite{yang2024dyfadet}  & ECCV 24    & Neck; Head          &     Dynamic Feature Aggregation               \\
            & CausalTAD~\cite{liu2024harnessing}  & ArXiv 24    & Neck          &     Causal Modeling               \\

\addlinespace[2pt]
\hdashline
\addlinespace[4pt]
\multirow{4}{*}{\makecell[l]{ Two-stage}}   & BMN~\cite{lin2019bmn}              & ICCV 19   & RoI          & Boundary Matching Mechanism               \\
            & GTAD~\cite{xu2020g}             & CVPR 20   & Neck           & Graph Convolutional Network \\
            & TSI~\cite{liu2020tsi}              & ACCV 20   & Loss           & Scale-Invariant Loss            \\
            & VSGN~\cite{zhao2021video}             & ICCV 21   & Neck        & Pyramid Cross-Scale Graph Network           \\
\addlinespace[2pt]
\hdashline
\addlinespace[4pt]
DETR  & TadTR~\cite{liu2022end}           & TIP 22    & Architecture          &  DETR + RoI Extraction                  \\
\midrule
\multicolumn{5}{c}{\textbf{End-to-End Approaches}} \\
\addlinespace[2pt]
\hdashline
\addlinespace[4pt]
Two-stage & AFSD~\cite{lin2021learning}             & CVPR 21   & RoI           &   Boundary Pooling                 \\
DETR      & E2E-TAD~\cite{liu2022empirical}           & CVPR 22   & Training           &  Empirical Study                  \\
Two-stage    & ETAD~\cite{liu2022etad}             & CVPRW 23  & Training           & Sequentialized Gradient Sampling                   \\
One | Two & Re$^2$TAL~\cite{zhao2023re2tal}           & CVPR 23   & Backbone           & Reversible Finetuning           \\
One-stage & AdaTAD~\cite{liu2024adatad}           & CVPR 24   & Backbone          & Adapter Tuning           \\
\bottomrule
\end{Tabular}
\label{tab:supported_methods}
\end{table}

OpenTAD supports \textbf{9 TAD datasets}, including widely used benchmarks such as ActivityNet-v1.3~\cite{caba2015activitynet}, THUMOS-14~\cite{jiang2014thumos}, HACS~\cite{zhao2019hacs}, and EPIC-Kitchens 100~\cite{damen2018scaling}, as well as recently published datasets like Ego4D-Moment Query~\cite{grauman2022ego4d} and FineAction~\cite{liu2022fineaction}. Beyond these single-label detection datasets, OpenTAD also supports multi-label datasets such as Multi-THUMOS~\cite{yeung2018every} and Charades~\cite{sigurdsson2016hollywood}. Additionally, OpenTAD extends support to audio-based action detection with datasets such as EPIC-Sounds~\cite{EPICSOUNDS2023}.
Thanks to OpenTAD’s modular design, switching between datasets is straightforward, enabling seamless adaptation of prior methods to all 9 TAD datasets, even if they were originally not implemented for them. In Table~\ref{tab:results_all_datasets} in the supplementary material, we report the detection performance of ActionFormer under the OpenTAD framework across all these datasets. Our results not only achieve comparable or superior performance on previously reported datasets such as THUMOS-14 and EPIC-Kitchens, but also establish new state-of-the-art results on newer datasets like FineAction.

\section{Experiments: A Comprehensive Study}

Despite numerous innovations in temporal action detection, evaluating their generalizability across different methods remains challenging. In this section, we leverage our unified framework, OpenTAD, with its modularized codebase to conduct a comprehensive study, fairly comparing various module designs and systematically analyzing how individual components, model sizes, overall architectures, and other factors affect TAD performance. 
We perform experiments with various TAD methods on the two most widely adopted datasets: ActivityNet-v1.3~\cite{caba2015activitynet} and THUMOS-14~\cite{jiang2014thumos}. We describe the evaluation metrics and implementation details in Sec.~\ref{sec:eval_implement_details} of the supplementary material.

\subsection{Which design for each component yields the best performance?}\label{sec:component_importance}

As described in Sec.~\ref{sec:categorization}, various design choices have been proposed for each component in the TAD literature. Among all the different variants, is there an absolute winner for each component?

\textbf{Neck.} This is an essential component present in almost every TAD method. As shown in Tab.~\ref{tab:supported_methods}, many methods propose innovative designs in the neck to enhance the temporal aggregation of video sequence features. Examples include the simple MaxPooling in TemporalMaxer~\cite{tang2023temporalmaxer}, the convolutional layer in BSN~\cite{lin2018bsn} and BMN~\cite{lin2019bmn}, the GCN layer in GTAD~\cite{xu2020g} and VSGN~\cite{zhao2021video}, the Transformer block in ActionFormer~\cite{zhang2022actionformer}, and the Mamba block in VideoMambaSuite~\cite{chen2024video}. 
We incorporate these neck designs into four different TAD methods: ActionFormer~\cite{zhang2022actionformer} and TriDet~\cite{shi2023tridet}, which produce multi-scale features, and GTAD~\cite{xu2020g} and BMN~\cite{lin2019bmn}, which produce single-scale features. We then compare their contributions to the final performance for each method, as shown in Tab.~\ref{tab:ablate_neck_thumos} and Tab.~\ref{tab:ablate_neck_anet} in the supplementary material.
\textit{Macro Block} in Column 1 refers to the highest-level repetitive building blocks, such as the following: (a) Transformer block, which contains a sequential modeling module and an MLP module, each with a residual connection; (b) Mamba block, which consists of a sequential modeling branch combined with a gated MLP branch; and (c) Block mix, which is a combination of the above two types of blocks. We partition Tab.~\ref{tab:ablate_neck_thumos} and Tab.~\ref{tab:ablate_neck_anet} into \textbf{four regions} based on the \textit{macro blocks}. Besides, \textit{Sequential Module} in Column 2 refers to the core sequential modeling module within the macro block, such as self-attention and SSM.

\begin{table}[t]
\centering
\caption{\textbf{Analysis of the neck design choices}, measured by average mAP(\%) on THUMOS-14. We ran all experiments 5 times and report their mean and standard deviation. The \textbf{4 macro-block regions} mean the following respectively. 
\textbf{Top region}: macro blocks with their original sequential modules are adopted as a whole; \textbf{Transformer block}:  self-attention modules in Transformer blocks are replaced with different sequential modules; \textbf{Mamba block}:  SSM modules in Mamba blocks are replaced with different sequential modules; \textbf{Bottom region}: a combination of two blocks.
}
\small
\setlength{\tabcolsep}{4.8pt}
\begin{tabular}{llccccc}
\toprule
\multicolumn{2}{c}{\textbf{Neck }} && \multicolumn{4}{c}{\textbf{Method}} \\
\addlinespace[2pt]
\cline{1-2} 
\cline{4-7} 
\addlinespace[4pt]
\textbf{Macro Block} &\multicolumn{1}{c}{\textbf{Sequential Module}} && {\textbf{ActionFormer}} & {\textbf{TriDet}} & {\textbf{BMN}} & {\textbf{GTAD}} \\
\midrule
Convolution Block& Convolution~\cite{lin2019bmn}    & & 52.58{\scriptsize $\pm$1.54} & 67.92{\scriptsize $\pm$0.22} & 48.43{\scriptsize $\pm$0.38} & 49.62{\scriptsize $\pm$0.22} \\         
GCN Block & Graph convolution~\cite{xu2020g}            & & 67.48{\scriptsize $\pm$0.22} & 68.32{\scriptsize $\pm$0.39} & 48.84{\scriptsize $\pm$0.26} & 50.36{\scriptsize $\pm$0.29}   \\
Transformer Block & Self-sttention~\cite{zhang2022actionformer} && 67.93{\scriptsize $\pm$0.19} & 68.58{\scriptsize $\pm$0.20} & \textbf{50.36}{\scriptsize $\pm$0.56} & \textbf{50.97}{\scriptsize $\pm$0.31}   \\
Mamba Block & SSM~\cite{chen2024video} && \textbf{68.28}{\scriptsize $\pm$0.48}  & \textbf{68.64}{\scriptsize $\pm$0.18} & 49.39{\scriptsize $\pm$0.33}        & 50.22{\scriptsize $\pm$0.35}  \\
\midrule
\multirow{5}{*}{\makecell[l]{Transformer Block}} 
&Convolution      && 68.11{\scriptsize $\pm$0.36} & 68.46{\scriptsize $\pm$0.52} & 48.73{\scriptsize $\pm$0.15} & 50.31{\scriptsize $\pm$0.16}  \\
&Graph convolution             & & 67.85{\scriptsize $\pm$0.26} & 68.51{\scriptsize $\pm$0.11} & 48.84{\scriptsize $\pm$0.26} & 50.33{\scriptsize $\pm$0.18}  \\
&Self-attention       && 67.93{\scriptsize $\pm$0.19} & 68.58{\scriptsize $\pm$0.20} & 50.36{\scriptsize $\pm$0.56} & 50.97{\scriptsize $\pm$0.31}  \\
&LSTM             && 68.10{\scriptsize $\pm$0.33} & 68.53{\scriptsize $\pm$0.27} & \textbf{51.66}{\scriptsize $\pm$0.35} & \textbf{53.31}{\scriptsize $\pm$0.27}  \\
& SSM             && \textbf{68.23}{\scriptsize $\pm$0.31} & \textbf{68.79}{\scriptsize $\pm$0.37}  & 49.23{\scriptsize $\pm$0.28} & 50.09{\scriptsize $\pm$0.35}  \\
\midrule
\multirow{5}{*}{\makecell[l]{Mamba  Block}} 
& Convolution     && 68.25{\scriptsize $\pm$0.34} & 68.61{\scriptsize $\pm$0.41} & 48.40{\scriptsize $\pm$0.20} & 50.11{\scriptsize $\pm$0.40} \\
& Graph convolution                && 68.02{\scriptsize $\pm$0.29} & 68.50{\scriptsize $\pm$0.45} & 48.73{\scriptsize $\pm$0.28} & 49.84{\scriptsize $\pm$0.33} \\
& Self-attention       && 68.24{\scriptsize $\pm$0.22} & 68.63{\scriptsize $\pm$0.26} & \textbf{49.57{\scriptsize $\pm$0.72}} & \textbf{50.89}{\scriptsize $\pm$0.70} \\
& LSTM            && 67.82{\scriptsize $\pm$0.29} & \textbf{68.85}{\scriptsize $\pm$0.56} & 49.20{\scriptsize $\pm$0.27} & 49.87{\scriptsize $\pm$0.31} \\
& SSM             && \textbf{68.28}{\scriptsize $\pm$0.48}  & 68.64{\scriptsize $\pm$0.18} & 49.39{\scriptsize $\pm$0.33}        & 50.22{\scriptsize $\pm$0.35}  \\
\midrule
Mamba + Mamba &SSM + SSM && 68.03{\scriptsize $\pm$0.38}  & 68.09{\scriptsize $\pm$0.87} & 49.66{\scriptsize $\pm$0.17} &  50.40{\scriptsize $\pm$0.48} \\
Transf. + Transf.&LSTM + Self-attention && 67.78{\scriptsize $\pm$0.31}  & 68.59{\scriptsize $\pm$0.62} & 50.91{\scriptsize $\pm$0.27} &  53.46{\scriptsize $\pm$0.56} \\
Mamba + Transf.&SSM + Self-attention  && 68.37{\scriptsize $\pm$0.22} & 68.41{\scriptsize $\pm$0.31} & 49.76{\scriptsize $\pm$0.46} & 50.60{\scriptsize $\pm$0.39}             \\
\textbf{Mamba + Transf.} &\textbf{SSM + LSTM} && \textbf{68.44}{\scriptsize $\pm$0.41} & \textbf{68.90}{\scriptsize $\pm$0.60} & \textbf{52.19}{\scriptsize $\pm$0.28} & \textbf{53.70}{\scriptsize $\pm$0.26} \\
\bottomrule
\end{tabular}
\label{tab:ablate_neck_thumos}
\end{table}

From the top region in Tab.~\ref{tab:ablate_neck_thumos}, we identify two winners: the Transformer block and the Mamba block, both of which feature more sophisticated macro architectures than the others. The Transformer block achieves the best performance for the single-scale methods BMN (50.36\%) and GTAD (50.97\%), while the Mamba block performs best for the multi-scale methods ActionFormer (68.28\%) and TriDet (68.64\%).
To further investigate which sequential modeling module within each macro architecture is the best choice, we replace the self-attention module in the Transformer block and the SSM module in the Mamba block with one of five sequential modeling modules: convolution, graph convolution, LSTM, self-attention, and SSM. We present the results in the \textit{Transformer Block} and \textit{Mamba Block} regions of Tab.~\ref{tab:ablate_neck_thumos} and Tab.~\ref{tab:ablate_neck_anet}. SSM demonstrates a clear advantage for the multi-scale methods ActionFormer and TriDet, while LSTM and self-attention exhibit superior performance for the single-scale methods BMN and GTAD. We attribute the strong performance of these three modules to their ability to perform temporal aggregation across the entire sequence, in contrast to the others.
It is also worth noting that when integrated into the Transformer block and the Mamba block, the performance of local sequential modeling modules—convolution ($52.58\% \rightarrow 68.11\%$ in ActionFormer) and GCN ($67.48\% \rightarrow 67.85\%$ in ActionFormer)—improves significantly compared to when they are used alone on THUMOS-14.

\textit{Can we identify a neck design that performs best across all methods?} Considering the outstanding performance of the Transformer and Mamba blocks, as well as the LSTM module, we propose a novel macro block that combines these two different blocks, as shown in the bottom region of Tab.~\ref{tab:ablate_neck_thumos} and Tab.~\ref{tab:ablate_neck_anet}. By mixing Mamba and Transformer macro blocks—where the self-attention in the Transformer block is replaced with LSTM—we achieve the highest performance across all four methods on both datasets, leading to an additional $0.8\%$ improvement on THUMOS-14 over the best-performing macro block with the optimal sequential module.

\textbf{RoI Extraction.} This component is utilized in two-stage methods and determines the quality of the proposal features used in Stage 2. Keypoint sampling, employed in PBRNet~\cite{liu2020progressive} and VSGN~\cite{zhao2021video}, extracts only the boundary and center features within a proposal. RoI Align~\cite{he2017mask} utilizes all features, while SGAlign~\cite{xu2020g}, which builds upon RoI Align, further enhances it by incorporating a GCN layer. Boundary matching, introduced in BMN~\cite{lin2019bmn}, densely extracts proposals based on every pair of start and end boundaries. We evaluate these four RoI strategies using three two-stage methods, with the results presented in Tab.~\ref{tab:ablate_RoI}.  

For both BMN and GTAD, boundary matching achieves the highest performance, while keypoint sampling yields the lowest. This is because both methods use a limited number of features due to their single-scale nature and the relatively small number of temporal length (e.g., 100 for ActivityNet). Given these constraints, the computational cost of using boundary matching to extract the largest number of proposals is acceptable. In contrast, for VSGN, which employs multi-scale feature maps and a larger number of input frames (e.g., 1280 frames for ActivityNet), applying dense boundary matching is infeasible, and keypoint sampling proves to be as effective as RoI Align. 
Therefore, the choice of RoI extraction method depends on specific properties of each approach, such as feature temporal scales. Since multi-scale, long-input methods like VSGN generally outperform conventional methods such as BMN and GTAD, boundary matching is generally not preferred.

\begin{table}[t]
\centering
\caption{\textbf{Analysis of the RoI extraction design choices}, measured by average mAP (\%) on two datasets. N/A means not applicable to the model. 
}
\small
\setlength{\tabcolsep}{5.3pt}
\begin{tabular}{lccccccc}
\toprule
&\multicolumn{3}{c}{\textbf{ActivityNet-v1.3}} & & \multicolumn{3}{c}{\textbf{THUMOS-14}} \\
\addlinespace[2pt]
\cline{2-4} 
\cline{6-8} 
\addlinespace[4pt]
\textbf{RoI Extraction} & \textbf{BMN} & {\textbf{GTAD}} & \textbf{VSGN} & & \textbf{BMN} &{\textbf{GTAD}} & \textbf{VSGN} \\
\midrule
Keypoint Sample~\cite{zhao2021video}    & 34.29{\scriptsize $\pm$0.07} & 32.68{\scriptsize $\pm$0.05} & \underline{36.71}{\scriptsize $\pm$0.13}  & & 43.36{\scriptsize $\pm$0.24} & 45.34{\scriptsize $\pm$0.59} & \underline{52.42}{\scriptsize $\pm$0.31}  \\
RoI Align~\cite{he2017mask}             & 36.06{\scriptsize $\pm$0.05} & 35.99{\scriptsize $\pm$0.13} & \textbf{36.74}{\scriptsize $\pm$0.02} & & \underline{49.48}{\scriptsize $\pm$0.67} & 49.40{\scriptsize $\pm$0.46} & \textbf{52.53}{\scriptsize $\pm$0.41} \\
SGAlign~\cite{xu2020g}                  & \underline{36.28}{\scriptsize $\pm$0.12} & \underline{36.24}{\scriptsize $\pm$0.04} & 36.66{\scriptsize $\pm$0.14} & & 49.00{\scriptsize $\pm$0.45} & \underline{50.36}{\scriptsize $\pm$0.29} & 52.38{\scriptsize $\pm$0.64} \\
{Boundary Match~\cite{lin2019bmn}}    & \textbf{36.44}{\scriptsize $\pm$0.07} & \textbf{36.34}{\scriptsize $\pm$0.07} & N/A & &\textbf{51.40}{\scriptsize $\pm$0.58} & \textbf{50.48}{\scriptsize $\pm$0.35} & N/A \\

\bottomrule
\end{tabular}
\label{tab:ablate_RoI}
\end{table}

\begin{table}[t]
\centering
\caption{\textbf{Analysis of the backbone design choices}, measured by average mAP (\%). Please note that all experiments use only the RGB modality, excluding the optical flow modality. Backbone models with * have been pre-pretrained using self-supervised learning. \textbf{Top region}: Resnet-based backbone models; \textbf{bottom region}: Transformer-based backbone models including VideoSwin and ViT architectures. `Param.': number of model parameters; K400: Kinetics-400; Top-1: top-1 accuracy (\%); `ActionF.': ActionFormer. We \colorbox{lightgray!50}{highlight} three backbones that don't have consistent action recognition and TAD performance.}
\small
\setlength{\tabcolsep}{1.3pt}
\begin{tabular}{lcccccccccccc}
\toprule
\multicolumn{1}{c}{\textbf{Backbone}} && \textbf{Param.} && \textbf{K400} &&\multicolumn{3}{c}{\textbf{ActivityNet-v1.3}} && \multicolumn{3}{c}{\textbf{THUMOS-14}} \\
\addlinespace[2pt]
\cline{3-3} 
\cline{5-5} 
\cline{7-9} 
\cline{11-13} 
\addlinespace[4pt]
\multicolumn{1}{c}{\textbf{Model}} && \textbf{(M)} && \textbf{Top-1} &&  {\textbf{ActionF.}} & {\textbf{GTAD}} & \textbf{VSGN} && {\textbf{ActionF.}} & {\textbf{GTAD}} & \textbf{VSGN} \\
\midrule
TSN-R50~\cite{TSN2016ECCV}  && 24 && 74.12  && 34.64{\scriptsize $\pm$0.07} & 34.09{\scriptsize $\pm$0.09} &33.36{\scriptsize $\pm$0.18} && 49.79{\scriptsize $\pm$0.37} & 35.18{\scriptsize $\pm$0.32} & 32.22{\scriptsize$\pm$0.79}\\
TSM-R50~\cite{lin2019tsm} && 24  && 75.12	&& 34.51{\scriptsize $\pm$0.08} & 33.70{\scriptsize $\pm$0.09} &34.35{\scriptsize $\pm$0.21}& &50.94{\scriptsize $\pm$0.30} & 34.59{\scriptsize $\pm$0.15} &35.09{\scriptsize$\pm$0.77} \\
R(2+1)D-R34~\cite{tran2018closer} && 64 && 75.46  && 35.06{\scriptsize $\pm$0.09} & 34.11{\scriptsize $\pm$0.06}&34.83{\scriptsize $\pm$0.22} && 48.39{\scriptsize $\pm$0.39} & 34.12{\scriptsize $\pm$0.23} & 39.08{\scriptsize$\pm$0.76} \\
SlowFast-R101~\cite{slowfast} && 63 && 78.65	&& 35.95{\scriptsize $\pm$0.12} & 35.28{\scriptsize $\pm$0.12} &35.98{\scriptsize $\pm$0.19} && 63.04{\scriptsize $\pm$0.33} & 47.08{\scriptsize $\pm$0.28}  & 50.54{\scriptsize$\pm$0.22} \\
\midrule
\rowcolor{lightgray!50} 
VideoSwin-S~\cite{vswin} && 50 && 80.54  && 35.60{\scriptsize $\pm$0.13} & 35.24{\scriptsize $\pm$0.10} & 32.85{\scriptsize $\pm$0.22} && 56.72{\scriptsize $\pm$0.70} & 38.99{\scriptsize $\pm$0.32} & 42.14{\scriptsize $\pm$0.56}   \\
\rowcolor{lightgray!50} 
VideoSwin-L~\cite{vswin} && 197 && 83.46  && 35.91{\scriptsize $\pm$0.03} & 35.73{\scriptsize $\pm$0.10} & 32.93{\scriptsize $\pm$0.32} && 58.78{\scriptsize $\pm$0.38} & 41.55{\scriptsize $\pm$0.16} & 45.66{\scriptsize$\pm$0.59}\\
\addlinespace[2pt]
\hdashline
\addlinespace[4pt]
\rowcolor{lightgray!50}  
MViTv2-L~\cite{li2022mvitv2}     && 213    && 85.40  && 35.96{\scriptsize $\pm$0.18} & 35.30{\scriptsize $\pm$0.12} &34.23{\scriptsize $\pm$0.18} && 47.62{\scriptsize $\pm$0.48} & 31.27{\scriptsize $\pm$0.51} & 41.78{\scriptsize$\pm$0.47}\\
VideoMAE-H*~\cite{tong2022videomae}   && 633    && 87.40  && 36.96{\scriptsize $\pm$0.07} & 36.57{\scriptsize $\pm$0.07} &35.07{\scriptsize $\pm$0.32} && 67.23{\scriptsize $\pm$0.32} & 48.16{\scriptsize $\pm$0.37}  & 53.93{\scriptsize$\pm$0.34}\\
VideoMAEv2-G*~\cite{wang2023videomae2}  && 1B    && 88.40  && - & - &- && 71.48{\scriptsize $\pm$0.12} & 54.16{\scriptsize $\pm$0.59} & 54.94{\scriptsize$\pm$0.64}\\
InternVideo2-6B*~\cite{wang2024internvideo2} && 6B  && \textbf{92.10}  && \textbf{38.95}{\scriptsize $\pm$0.13} & \textbf{37.35}{\scriptsize $\pm$0.12} &\textbf{38.63}{\scriptsize $\pm$0.09} && \textbf{72.36}{\scriptsize $\pm$0.13}  & \textbf{56.32}{\scriptsize $\pm$0.16} & \textbf{58.16}{\scriptsize $\pm$0.66}\\
\bottomrule
\end{tabular}
\vspace{-5pt}
\label{tab:ablate_backbone}
\end{table}

\textbf{Backbone}. It produces video features, which are to be further processed by the subsequent components of the TAD framework (e.g., the neck and the heads) to make predictions. Therefore, the quality of the video features and the backbone is critical, significantly influencing the detection performance. A backbone network is usually pretrained on a larger-scale action recognition dataset such as Kinetics~\cite{zisserman2017kinetics} before being trained for TAD, and example backbones are TSN~\cite{TSN2016ECCV}, SlowFast~\cite{slowfast}, and VideoSwin~\cite{vswin}. Recently,  ViT~\cite{dosovitskiy2021image} is adopted as video backbone network, pre-pretrained with self-supervised learning, multi-modal joint learning, or other methods (e.g., VideoMAE~\cite{tong2022videomae}, InternVideo2~\cite{wang2024internvideo2}), and has shown remarkable performance on TAD. In Tab.~\ref{tab:ablate_backbone}, we present experiments across three TAD methods and two datasets,  evaluating different backbone models. These models vary in architecture, including ResNet-based and Transformer-based designs, and in pretraining strategies, including action recognition pretraining and additional pre-pretraining.

In Tab.~\ref{tab:ablate_backbone}, the backbone model sizes are arranged in non-decreasing order from top to bottom within each group: ResNet-based backbones in the top region and Transformer-based backbones in the bottom region.
We observe that action recognition performance on Kinetics-400 consistently improves as model sizes increase (Column 2), and that all the Transformer-based models perform better than the ResNet-based models. However, this trend does not always hold for TAD.  Though VideoSwin-S~\cite{vswin} and MViTv2-L~\cite{li2022mvitv2}   have significantly higher action recognition accuracy than SlowFast-101~\cite{slowfast}, their mAPs are lower for most TAD experiments.  Similarly, while MViTv2-L outperforms VideoSwin-L on Kinetics, it performs worse on THUMOS-14 for TAD. We attribute these inconsistencies across tasks to the architectural differences and suggest that ViT-based models are more effective for TAD. Notably, the largest ViT-based model, InternVideo2-6B~\cite{wang2024internvideo2} achieves the highest performance across all methods and datasets.

\subsection{Is the optional Stage 2 necessary?}
Stage 2 is optional in the TAD framework, as it is only used in two-stage methods. While one-stage methods were originally introduced to improve efficiency in object detection~\cite{sermanet2013overfeat}, the question remains: is Stage 2 necessary for achieving higher accuracy?

To answer this question, we compare the performance of methods with and without Stage 2 across three datasets in Tab.~\ref{tab:ablate_category_stage2}, including originally one-stage, two-stage, and DETR-based methods. For one-stage methods, we incorporate Stage 2 by using RoI Align~\cite{he2017mask} for RoI extraction and applying the same classification and regression losses as in Stage 1. We define positive samples as proposals with $\textrm{tIoU}>0.7$ relative to the ground-truth actions. For two-stage methods, we entirely remove Stage 2, using Stage 1 predictions as the final output for ablation. Additionally, since the DETR-based method TadTR~\cite{liu2022end} includes a second-stage refinement, we also evaluate its performance without this stage for comparison. As shown in Tab.\ref{tab:ablate_category_stage2}, adding Stage 2 improves the performance of the one-stage method ActionFormer\cite{zhang2022actionformer}, while removing Stage 2 reduces the performance of the two-stage method VSGN~\cite{zhao2021video} and the DETR-based method TadTR.  These results indicate that Stage 2 can further enhance one-stage methods, and is essential for two-stage methods.

\begin{table}[t]
\centering
\caption{\textbf{Ablation study of Stage 2} on one-stage, two-stage, and DETR-based methods, measured by average mAP (\%). N/A means training details are not provided in the original paper or code.}
\small
\begin{tabular}{llccc}
\toprule
\textbf{Category} & \textbf{Method} &\textbf{ActivityNet-v1.3} & \textbf{THUMOS-14}   & \textbf{HACS} \\
\midrule
\multirow{2}{*}{{One-stage}}      
& ActionFormer~\cite{zhang2022actionformer}              & 37.00{\scriptsize $\pm$0.05} & 67.93{\scriptsize $\pm$0.19} & 37.69{\scriptsize $\pm$0.05}\\
& ActionFormer + Stage 2    & \textbf{37.29}{\scriptsize $\pm$0.03} & \textbf{68.61}{\scriptsize $\pm$0.29} & \textbf{38.25}{\scriptsize $\pm$0.12}\\
\midrule
\multirow{2}{*}{{Two-stage}}
& VSGN - Stage 2 & 36.18{\scriptsize $\pm$0.07} & 49.71{\scriptsize $\pm$0.38} & 36.02{\scriptsize $\pm$0.20} \\
& VSGN~\cite{zhao2021video}           & \textbf{36.71}{\scriptsize $\pm$0.13}  & \textbf{52.42}{\scriptsize $\pm$0.31}  & \textbf{37.19}{\scriptsize $\pm$0.12} \\
\midrule
\multirow{2}{*}{{DETR-based}} & TadTR - Stage 2 & N/A& 52.75{\scriptsize $\pm$0.55} & N/A \\
& TadTR~\cite{liu2022end}  & N/A & \textbf{54.42}{\scriptsize $\pm$0.72} & N/A\\
\bottomrule
\end{tabular}
\label{tab:ablate_category_stage2}
\end{table}

\subsection{What makes an optimal detector?}

Through the above analysis within our unified OpenTAD framework, we have identified the optimal design choices for the neck, RoI extraction, and backbone components. This raises an intriguing question: can we further enhance existing TAD methods by applying these winning techniques?

In Tab.~\ref{tab:apply_all_winners}, we use ActionFormer as the baseline (Row 1) and progressively apply the winning techniques—excluding backbone modifications—across different components (Rows 2–5). We evaluate the impact of these changes on two datasets. As expected, performance improves progressively as more techniques are applied, culminating in a model with mAP improvements of 0.62/0.88 over the original. Next, we separately replace the backbone with the winning InternVideo2 model and then apply all other techniques (Rows 6–7). While backbone replacement alone substantially boosts performance, incorporating additional modifications further enhances the results, achieving an overall improvement of 0.56/1.06 mAP.

In Tab.~\ref{tab:sota_results}, we apply all the modifications—except for the backbone replacement—to five different TAD methods across two datasets, consistently improving their performance. Notably, for the recent method VideoMambaSuite~\cite{chen2024video}, which utilizes the powerful InternVideo2 backbone, our modifications further enhance its results, achieving state-of-the-art performance on both ActivityNet-v1.3 and THUMOS-14.

\begin{table}[t]
\centering
\caption{\textbf{Progressive modification of components in ActionFormer~\cite{zhang2022actionformer}} , performance measured by average mAP (\%). \textbf{Row 1}: baseline ActionFormer with TSP features for ActivityNet-v1.3 and I3D features for THUMOS-14; \textbf{Top region}: progressive adding or removing a component or a technique; \textbf{Bottom region}: replacing the backbone and applying all the modifications in the top region.}
\small
\begin{tabular}{lcccc}
\toprule
\textbf{Setting} & \textbf{ActivityNet-v1.3} & \textbf{THUMOS-14} \\
\midrule
ActionFormer (TSP backbone | I3D backbone): baseline      & 37.00{\scriptsize $\pm$0.05} & 67.93{\scriptsize $\pm$0.19}   \\
\midrule
- Remove the transformer block & 36.04{\scriptsize $\pm$0.33} & 67.52{\scriptsize $\pm$0.36}    \\
+ Add block mix & 37.50{\scriptsize $\pm$0.07}  & 68.44{\scriptsize $\pm$0.41} \\
+ Add the second stage to refine proposals & 37.59{\scriptsize $\pm$0.08} & 68.63{\scriptsize $\pm$0.33} \\
+ Use input channel dropout, longer training epochs & \textbf{37.62}{\scriptsize $\pm$0.02} & \textbf{68.81}{\scriptsize $\pm$0.43}  \\
\midrule 
Replace the backbone with InternVideo2  
& 38.95{\scriptsize $\pm$0.13} & 72.36{\scriptsize $\pm$0.13}   \\
+ All the above modifications & \textbf{39.51}{\scriptsize $\pm$0.09} & \textbf{73.42}{\scriptsize $\pm$0.40} \\
\bottomrule
\end{tabular}
\label{tab:apply_all_winners}
\end{table}

\begin{table}[t]
\centering
\caption{\textbf{Results on ActivityNet-v1.3 and THUMOS-14}, measured by mAP (\%) at different tIoU thresholds. TSP~\cite{alwassel2021tsp} features are used for ActivityNet, and I3D~\cite{carreira2017quo} features are used for THUMOS for all methods except VideoMambaSuite, which uses the InternVideo2~\cite{wang2024internvideo2} features for both datasets.}
\small
\setlength{\tabcolsep}{4.7pt}
\begin{tabular}{lcccacccccca}
\toprule
\multirow{2}{*}{\textbf{Method}} & \multicolumn{4}{c}{\textbf{ActivityNet-v1.3}} & \multicolumn{6}{c}{\textbf{THUMOS-14}} \\
\addlinespace[2pt]
\cline{2-5} 
\cline{7-12} 
\addlinespace[4pt]
& \textbf{0.5} & \textbf{0.75} & \textbf{0.95} & \textbf{Avg.}  & & \textbf{0.3} & \textbf{0.4} & \textbf{0.5} & \textbf{0.6} & \textbf{0.7} & \textbf{Avg.} \\
\midrule
GTAD~\cite{xu2020g}         & 52.33 & 37.58 & 8.42 & 36.20 & & 64.35 & 59.07 & 51.76 & 42.65 & 31.66 & 49.70 \\
\textbf{GTAD +}             & 52.46 & 37.86 & 9.55 & \textbf{36.67} & & 67.08 & 62.87 & 56.81 & 46.68 & 35.74 & \textbf{53.83} \\  
\addlinespace[2pt]
\hdashline
\addlinespace[4pt]
BMN~\cite{lin2019bmn}       & 52.90 & 37.30 & 9.67 & 36.40 & & 64.99 & 60.70 & 54.54 & 44.11 & 34.16 & 51.80 \\
\textbf{BMN +}              & 53.82 & 38.14 & 9.92 & \textbf{37.04} & & 66.82 & 61.94 & 55.22 & 45.16 & 34.11 & \textbf{52.65} \\
\addlinespace[2pt]
\hdashline
\addlinespace[4pt]
ActionFormer~\cite{zhang2022actionformer}  & 55.08 & 38.27 & 8.91 & 37.07 & & 83.78 & 80.06 & 73.16 & 60.46 & 44.72 & 68.44  \\
\textbf{ActionFormer +} & 55.76 & 39.13 & 7.43 & \textbf{37.68} & & 83.05 & 79.20 & 72.20 & 62.73 & 47.25 & \textbf{68.89}\\
\addlinespace[2pt]
\hdashline
\addlinespace[4pt]
TriDet~\cite{shi2023tridet} & 54.84 & 37.46 & 7.98 & 36.51 & & 84.46 & 81.05 & 73.41 & 62.58 & 46.51 & 69.60 \\
\textbf{TriDet +} & 55.49 & 38.65 & 8.95 & \textbf{37.47} & & 84.49 & 80.77 & 73.71 & 62.22 & 47.16 & \textbf{69.67} \\
\addlinespace[2pt]
\hdashline
\addlinespace[4pt]
VideoMambaSuite~\cite{chen2024video} & 63.13 & 44.36 & 10.36 & 42.80 && 87.30 & 82.95 & 77.17 & 67.06 & 51.74 & 73.24 \\
\textbf{VideoMambaSuite +} & 63.22 & 45.01 & 8.79 & \textbf{43.02} && 86.79 & 83.11 & 77.54 & 67.21 & 53.66 & \textbf{73.66} \\
\bottomrule
\end{tabular}
\label{tab:sota_results}
\end{table}
\section{Related Works}

\subsection{TAD Frameworks}

The absence of a standardized framework for temporal action detection has historically led researchers to either develop methods from scratch or build upon specific open-source implementations. While some methods have provided partial foundations, they remain limited in scope. For example, GTAD~\cite{xu2020g} was implemented on top of BMN~\cite{lin2019bmn} and was later adopted by subsequent works such as TSI~\cite{liu2020tsi} and BC-GNN~\cite{bai2020boundary}. However, these methods are tailored to specific designs and lack the flexibility to accommodate broader innovations. Additionally, adapting a feature-based method (e.g., \cite{zhao2021video}) to an end-to-end training paradigm (e.g., \cite{zhao2023re2tal}) often requires extensive and tedious code modifications, such as restructuring feature extraction and training components. This lack of interoperability hinders efficient experimentation and fair benchmarking.

Although several works have proposed distinct ``frameworks" for TAD, they primarily focus on specific paradigms rather than providing a truly unified solution. Examples include the \textit{one-stage framework} in TriDet~\cite{shi2023tridet}, the \textit{multi-level cross-scale framework} in VSGN~\cite{zhao2021video}, the \textit{end-to-end detection framework} in TadTR~\cite{liu2022end}, and the \textit{efficient end-to-end framework} in AdaTAD~\cite{liu2024adatad}. These frameworks are designed independently, making cross-method comparison and integration challenging. Some methods, such as BMN~\cite{lin2019bmn} and TSI~\cite{liu2020tsi}, describe their approaches as ``unified frameworks," but they primarily refer to the joint training of multiple components within a single model rather than a framework that accommodates diverse methodologies.

To address these limitations, OpenTAD introduces the first truly unified framework for TAD, integrating a broad range of approaches—including one-stage, two-stage, DETR-based, and end-to-end methods—within a single, cohesive implementation. By supporting both feature-based and end-to-end learning paradigms, OpenTAD simplifies the development, adaptation, and evaluation of new methods, ensuring a standardized and extensible platform for future research.

\subsection{TAD Surveys}

Several survey papers on temporal action detection provide comprehensive overviews of various methods~\cite{wang2023temporal, hu2024overview, vahdani2022deep}. These reviews categorize TAD approaches based on the original modularization described in each paper and directly compare reported performance across different methods. However, they do not account for inconsistencies in experimental configurations, such as data resolution, preprocessing, and post-processing, which can lead to unfair comparisons and obscure the true effectiveness of individual techniques.

In contrast, OpenTAD offers a unified framework that systematically re-modularizes and reimplements existing methods to ensure consistency across different architectures and training pipelines. This standardization enables faithful comparison and rigorous analysis, providing deeper insights into the impact of specific design choices. Unlike previous survey papers, which evaluate methods based on heterogeneous configurations, OpenTAD facilitates direct, controlled comparisons under a unified setting. Furthermore, while prior reviews independently classify TAD methods into categories, OpenTAD integrates diverse approaches—including one-stage, two-stage, DETR-based, and end-to-end methods—into a single, cohesive implementation, enabling a more comprehensive and adaptable benchmarking platform.
\section{Conclusion}

In this work, we introduced OpenTAD, a unified framework that standardizes and modularizes temporal action detection across diverse methods and datasets, providing the most extensive TAD codebase available, to the best of our knowledge. 
OpenTAD facilitates fair comparisons across different methods and enables seamless plug-in implementations for new innovations within its modular components. 
Leveraging OpenTAD, we conducted a comprehensive analysis and identified key factors that significantly impact TAD performance, such as second-stage processing and neck architectures. Based on these insights, we developed a new state-of-the-art TAD method. 
We invite the research community to engage with OpenTAD, which is publicly available along with all associated resources, to drive further innovation in video understanding. Through collaborative efforts, we envision OpenTAD advancing not only action detection but also broader video understanding tasks.

{
\small
\bibliographystyle{plain}
\bibliography{main}
}

\appendix
\newpage
\section*{Supplementary Material}

\section{Evaluation Protocol and Implementation Details} \label{sec:eval_implement_details}

\textbf{Evaluation Protocol}. Although mean Average Precision (mAP) is the standard evaluation metric for TAD, inconsistencies in evaluation code, ground-truth annotations, and tIoU thresholds across previous methods have led to difficulties in fair comparisons. In the OpenTAD framework, we standardize the evaluation protocol across datasets and methods, ensuring consistency. We report the mean and standard deviation of performance metrics over five different random seeds to account for variability in training.

\noindent \textbf{Implementation Details}.
\label{supp:implementation_details}
OpenTAD is implemented using PyTorch 2.0 and runs on a single NVIDIA A100 GPU. We adhere to each method's original hyperparameter settings, including learning rate, number of training epochs, and other relevant parameters. Unless otherwise stated, we use pre-extracted TSP features on ActivityNet and I3D features on THUMOS for our ablation studies. Batch sizes and optimizer configurations are kept consistent with each method’s original implementation.

\section{Introduction and Categorization of Implemented Methods}
\label{supp:category}

OpenTAD implements 16 representative temporal action detection methods. To align each paper’s original design with OpenTAD’s modular framework, we categorize the methods based on their sub-components, such as neck, dense head, RoI, loss functions, and more. Table~\ref{tab:mapping_to_opentad} provides a detailed mapping of these components. We classify the methods into four categories: one-stage, two-stage, DETR-based, and end-to-end methods. From this table, we can clearly see that the primary difference between one-stage and two-stage methods is that the latter includes an additional step involving RoI extraction to further refine the candidate actions. 

For reproducibility, we re-implemented all 16 methods on the ActivityNet and THUMOS datasets, with results reported in Table~\ref{tab:re-implementation_results}. Our results closely match the original papers, and in some cases, such as BMN, we achieve significantly better performance. Additionally, OpenTAD also supports 9 widely used temporal action detection datasets: ActivityNet-v1.3~\cite{caba2015activitynet}, THUMOS-14~\cite{jiang2014thumos}, EPIC-Kitchens 100~\cite{damen2018scaling}, Ego4D-MQ~\cite{grauman2022ego4d}, HACS~\cite{zhao2019hacs}, Multi-THUMOS~\cite{yeung2018every}, Charades~\cite{sigurdsson2016hollywood}, FineAction~\cite{liu2022fineaction}, and EPIC-Sounds~\cite{EPICSOUNDS2023}. We benchmarked ActionFormer as the base model across all datasets, with results presented in Table~\ref{tab:results_all_datasets}.

\begin{table}[h]
\centering
\small
\caption{\textbf{Re-implemented results of different methods in OpenTAD} in terms of mAP (\%).   N/A means not provided in the paper or released code. For THUMOS-14, previous papers usually reported 5 numbers from mAP at tIoU=\{0.1, 0.2, 0.3, 0.4, 0.5, 0.6, 0.7\}, and some reported average mAP values, which mean differently across papers. Here, we standardize average mAP as the average of the mAP values at tIoU=\{0.3, 0.4, 0.5, 0.6, 0.7\}, compute this number based on the reported mAP at these 5 tIoUs. For ActivityNet, we compute the average mAP at tIoU=\{0.5:0.95:0.05\}.}
    \setlength{\tabcolsep}{3.2pt}
    \begin{tabular}{llcacacca}
        \toprule
        \multirow{3}{*}{\textbf{Method}} & \multirow{3}{*}{\textbf{Backbone}} & \multicolumn{4}{c}{\textbf{THUMOS-14}} && \multicolumn{2}{c}{\textbf{ActivityNet-v1.3}} \\
        \addlinespace[2pt]
        \cline{3-6} 
        \cline{8-9}  
        \addlinespace[4pt]
        &  & Original & OpenTAD  & Original  & OpenTAD  &&  Original & OpenTAD \\
        &  & tIoU=0.5 & tIoU=0.5  &  Average & Average && Average & Average \\
        \midrule
        ActionFormer~\cite{zhang2022actionformer} & I3D | TSP     & 71.00  & 73.16 & 66.84 & 68.44 && 36.60  & 37.07 \\
        TemporalMaxer~\cite{tang2023temporalmaxer} & I3D | TSP    & 71.80  & 71.66 & 67.70 & 68.33 && N/A   &  N/A  \\
        TriDet~\cite{shi2023tridet} & I3D | TSP           & 72.90  & 73.41 & 69.28 & 69.60 && 36.80 & 36.51 \\
        CausalTAD~\cite{liu2024harnessing} & I3D | TSP    & 73.57  & 73.57 & 69.75 & 69.75 && 37.46  &  37.46  \\
        DyFADet~\cite{yang2024dyfadet} & \tiny{VideoMAEv2-G | TSP}    & 73.70  & 76.32 & 70.50 & 71.70 && 38.50  &  38.62  \\
        VideoMambaSuite~\cite{chen2024video} & InternVideo2         & 76.90 & 77.17 & 72.72 & 73.24 && 42.02 & 42.80 \\
        \midrule
        BMN~\cite{lin2019bmn} & TSN                         & 38.80  & 47.56 & 38.48 & 46.19 && 33.85 & 34.21 \\
        GTAD~\cite{xu2020g} & TSN                           & 43.04 & 48.50 & 41.41 & 46.49 && 34.09 & 34.18 \\
        TSI~\cite{liu2020tsi} & TSN | TSP                     & 42.60  & 46.14 & 42.26 & 44.75 && 35.24 & 35.36 \\
        VSGN~\cite{zhao2021video} & TSN | TSP                 & 45.52 & 49.37 & 43.37 &47.25  && 35.94 & 36.89 \\
        \midrule
        TadTR~\cite{liu2022end} & I3D | TSP & 60.10 & 59.00 & 56.68 & 56.23 && 36.75 & N/A \\
        \midrule
        AFSD~\cite{lin2021learning} & I3D | TSP           & 55.50  & 60.16 & 52.00 & 55.96 && 34.40 &  36.10 \\
        E2E-TAD~\cite{liu2022empirical} & I3D           & 47.00  & 59.00 & 45.08 & 56.23 &&  N/A   &  N/A  \\
        ETAD~\cite{liu2022etad} & R(2+1)D               & 56.17 & 58.23 & 54.66 & 55.56 && 38.25 & 38.76 \\
        Re$^2$TAL~\cite{zhao2023re2tal} & \tiny{Re$^2$SlowFast-101}  & 64.90  & 74.27 & 61.52 & 70.19 && 37.01 & 37.55 \\
        AdaTAD~\cite{liu2024adatad} & \tiny{VideoMAEv2-G}      & 80.90  & 81.24 & 76.88 & 77.07 && 41.93 & 41.85 \\
        \bottomrule
    \end{tabular}
    \label{tab:re-implementation_results}
\end{table}

\begin{table}[h]
\centering
\caption{\textbf{Detection performance on all 9 supported datasetsusing ActionFormer~\cite{zhang2022actionformer},} measured by average mAP(\%). N/A. means not implemented in ActionFormer's original codebase.}
\footnotesize
\begin{tabular}{@{}lccccc}
\toprule
\textbf{Method}       & \textbf{THUMOS-14}     & \textbf{ActivityNet-v1.3}   & \textbf{EPIC-Kitchens}  & \textbf{Ego4D-MQ} & \textbf{HACS} \\
\midrule
Original     & 66.83      & 36.56         & 21.88 | 23.51   & 23.29 & N/A  \\
OpenTAD      & 68.44      & 37.07         & 22.33 | 24.93   & 25.57  & 37.71   \\
\midrule
\textbf{Method}     & \textbf{Multi-THUMOS}  & \textbf{Charades} & \textbf{FineAction} & \textbf{EPIC-Sounds}\\
\midrule
Original       & N/A          & N/A  & N/A  & N/A  \\
OpenTAD       & 39.18         & 19.39  & 19.62 & 13.89  \\
\bottomrule
\end{tabular}
\vspace{-5pt}
\label{tab:results_all_datasets}
\end{table}

\section{Benchmark Results}
\label{supp:benchmark_result}

In this section, we present the benchmark results for all implemented methods within the OpenTAD framework. For ActivityNet, we use TSP features with standardized evaluation annotations across all methods. For THUMOS, we adopt two-stream I3D features while ensuring consistency in evaluation annotations. The results are reported in Table~\ref{tab:benchmark_anet} and Table~\ref{tab:benchmark_thumos}.

From our benchmarks, we observe that one-stage detection methods have emerged as the preferred choice for both datasets. On THUMOS, two-stage methods such as GTAD and BMN, which rely on external classifiers for action classification, generally underperform compared to recent one-stage methods. However, on ActivityNet, where both one-stage and two-stage methods leverage external video-level classification results, two-stage methods with cascaded proposal refinements still achieve slightly better performance than their one-stage counterparts.

\begin{table}[h]
\centering
\small
\caption{\textbf{Benchmarking results on ActivityNet-v1.3 with TSP feature} in terms of mAP (\%).  }
\begin{tabular}{lrccca}
\toprule
\textbf{Method} & \textbf{\#Param. (M)}  & \textbf{0.5} & \textbf{0.75} & \textbf{0.95} & \textbf{Avg. mAP} 	\\
\midrule
TSI~\cite{liu2020tsi}                       & 4.54  & 52.44 & 35.57 & 9.80 & 35.36 \\
TemporalMaxer~\cite{tang2023temporalmaxer}  & 1.38  & 54.59 & 37.13 & 7.11 & 36.03 \\
AFSD~\cite{lin2021learning}                 & 13.41  & 54.44 & 36.72 & 8.69 & 36.10 \\
GTAD~\cite{xu2020g}                         & 5.58  & 52.33 & 37.58 & 8.42 & 36.20 \\
BMN~\cite{lin2019bmn}                       & 2.80  & 52.90 & 37.30 & 9.67 & 36.40 \\
TriDet~\cite{shi2023tridet}                 & 12.81 & 54.84 & 37.46 & 7.98 & 36.51 \\
VSGN~\cite{zhao2021video}                   & 6.50 & 54.80 & 37.35 & 9.80 & 36.89 \\
ActionFormer~\cite{zhang2022actionformer}   & 6.94  & 55.08 & 38.27 & 8.91 & 37.07 \\
VideoMambaSuite~\cite{chen2024video}        & 4.30 & 55.61 & 38.49 & 9.18 & 37.45 \\
CausalTAD~\cite{liu2024harnessing}          & 12.75  & 55.62 & 38.51 & 9.40 & 37.46 \\
ETAD~\cite{liu2022etad}                     & 8.52  & 54.91 & 38.98 & 9.09 & 37.73 \\
        \bottomrule
    \end{tabular}
    \label{tab:benchmark_anet}
\end{table}

\begin{table}[h]
\centering
\caption{\textbf{Benchmarking results on THUMOS14 with I3D feature} in terms of mAP (\%).  }
\small
\setlength{\tabcolsep}{4.pt}
\begin{tabular}{lrccccca}
\toprule
\textbf{Method} & \textbf{\#Param.(M)}  & \textbf{0.3} & \textbf{0.4} & \textbf{0.5} & \textbf{0.6} & \textbf{0.7} & \textbf{Avg. mAP} 	\\
\midrule
TSI~\cite{liu2020tsi}                       & 4.84 & 62.56 & 57.00 & 50.22 & 40.18 & 30.17 & 48.03  \\
GTAD~\cite{xu2020g}                         & 6.14  & 64.35 & 59.07 & 51.76 & 42.65 & 31.66 & 49.70  \\
BMN~\cite{lin2019bmn}                       & 3.10  & 64.99 & 60.70 & 54.54 & 44.11 & 34.16 & 51.80  \\
VSGN~\cite{zhao2021video}                   & 8.37  & 68.25 & 62.46 & 54.99 & 44.07 & 32.36 & 52.43  \\
ETAD~\cite{liu2022etad}                     & 5.37  & 67.74 & 64.22 & 58.23 & 49.19 & 38.41 & 55.56  \\
AFSD~\cite{lin2021learning}                 & 14.24  & 73.20 & 68.45 & 60.16 & 46.74 & 31.24 & 55.96  \\
TadTR~\cite{liu2022end}                     & 8.66  & 71.90 & 67.29 & 59.00 & 48.34 & 34.61 & 56.23 \\
TemporalMaxer~\cite{tang2023temporalmaxer}  & 7.12  & 83.17 & 79.09 & 71.66 & 61.72 & 46.00 & 68.33 \\
ActionFormer~\cite{zhang2022actionformer}   & 29.25  & 83.78 & 80.06 & 73.16 & 60.46 & 44.72 & 68.44 \\
VideoMambaSuite~\cite{chen2024video}        & 18.57  & 84.33 & 80.60 & 74.19 & 61.99 & 46.71 & 69.57 \\
TriDet~\cite{shi2023tridet}                 & 15.99  & 84.46 & 81.05 & 73.41 & 62.58 & 46.51 & 69.60 \\
CausalTAD~\cite{liu2024harnessing}          & 52.11  & 84.43 & 80.75 & 73.57 & 62.70 & 47.33 & 69.75 \\
        \bottomrule
    \end{tabular}
    \label{tab:benchmark_thumos}
\end{table}

\section{Supplementary Experiments}
\label{supp:more_experiments}

\subsection{Ablation Study on Neck Design in the THUMOS Dataset}

To further examine the impact of neck design, we conduct an ablation study on the ActivityNet dataset, with results presented in Table~\ref{tab:ablate_neck_anet}. These findings align with those reported in Table~\ref{tab:ablate_neck_thumos} of the main paper. 
Our experiments show that LSTM achieves performance comparable to or even better than the SSM module in BMN and GTAD. Furthermore, by integrating both designs, we obtain the best overall performance across all four evaluated methods.

\begin{table}[t]
\centering
\caption{\textbf{Analysis of the neck design choices}, measured by average mAP(\%) on ActivityNet-v1.3. The \textbf{4 macro-block regions} mean the following respectively. \textbf{Top region}: macro blocks with their original sequential modules are adopted as a whole; \textbf{Transformer Block}:  self-attention modules in Transformer blocks are replaced with different sequential modules; \textbf{Mamba Block}:  SSM modules in Mamba blocks are replaced with different sequential modules; \textbf{Bottom region}: a combination of two blocks.
Note that in Row 1,  BMN and GTAD directly use identity mapping since they don't downscale temporally. TSP~\cite{alwassel2021tsp} features are used.}
\small
\begin{tabular}{llccccc}
\toprule
\multicolumn{2}{c}{\textbf{Neck }} && \multicolumn{4}{c}{\textbf{Method}} \\
\addlinespace[2pt]
\cline{1-2} 
\cline{4-7} 
\addlinespace[4pt]
\textbf{Macro Block} &\multicolumn{1}{c}{\textbf{Sequential Module}} && {\textbf{ActionFormer}} & {\textbf{TriDet}} & {\textbf{BMN}} & {\textbf{GTAD}} \\
\midrule
Convolution Block& Convolution    && 36.88{\scriptsize $\pm$0.03} & 36.87{\scriptsize $\pm$0.02} & 36.44{\scriptsize $\pm$0.05} & 36.36{\scriptsize $\pm$0.09} \\      
GCN Block& Graph convolution           && 37.03{\scriptsize $\pm$0.04} & 37.00{\scriptsize $\pm$0.05} & 36.41{\scriptsize $\pm$0.02} & 36.24{\scriptsize $\pm$0.04} \\
Transformer Block& Self-attention&& 37.00{\scriptsize $\pm$0.05} & 36.93{\scriptsize $\pm$0.09} & \textbf{36.50}{\scriptsize $\pm$0.03} & 36.31{\scriptsize $\pm$0.06} \\
Mamba Block& SSM  && \textbf{37.40}{\scriptsize $\pm$0.03} & \textbf{37.33}{\scriptsize $\pm$0.07} &36.36{\scriptsize $\pm$0.05}  & \textbf{36.36}{\scriptsize $\pm$0.06} \\
\midrule
\multirow{5}{*}{\makecell[l]{Transformer  Block}} 
&Convolution     && 36.99{\scriptsize $\pm$0.02} & 36.93{\scriptsize $\pm$0.09} & 36.40{\scriptsize $\pm$0.07} & 36.31{\scriptsize $\pm$0.07}   \\
&Graph convolution             && 37.15{\scriptsize $\pm$0.03} & 36.98{\scriptsize $\pm$0.07} & 36.41{\scriptsize $\pm$0.02} & 36.13{\scriptsize $\pm$0.08} \\
&Self-attention       && 37.00{\scriptsize $\pm$0.05} & 36.93{\scriptsize $\pm$0.09} & 36.50{\scriptsize $\pm$0.03} & 36.31{\scriptsize $\pm$0.06} \\
&LSTM            && 37.18{\scriptsize $\pm$0.04} & 36.95{\scriptsize $\pm$0.07} & \textbf{36.87}{\scriptsize $\pm$0.09} & \textbf{36.47}{\scriptsize $\pm$0.12}  \\
& SSM            && \textbf{37.40}{\scriptsize $\pm$0.04} & \textbf{37.33}{\scriptsize $\pm$0.07} & 36.58{\scriptsize $\pm$0.23} & 36.21{\scriptsize $\pm$0.08}  \\
\midrule
\multirow{5}{*}{\makecell[l]{Mamba  Block}} 
& Convolution    && 37.07{\scriptsize $\pm$0.03}  & 36.97{\scriptsize $\pm$0.07} & 36.40{\scriptsize $\pm$0.05} & 36.17{\scriptsize $\pm$0.08}\\
& Graph convolution            && 37.13{\scriptsize $\pm$0.07}  & 37.06{\scriptsize $\pm$0.04} & 36.33{\scriptsize $\pm$0.05} & 36.16{\scriptsize $\pm$0.09} \\
& Self-attention      && 36.21{\scriptsize $\pm$0.22}  & 36.33{\scriptsize $\pm$0.14} & 36.35{\scriptsize $\pm$0.07} & 36.12{\scriptsize $\pm$0.04} \\
& LSTM           && 36.95{\scriptsize $\pm$0.11}  & 36.57{\scriptsize $\pm$0.07} & \textbf{36.42}{\scriptsize $\pm$0.08} & \textbf{36.23}{\scriptsize $\pm$0.05} \\
& SSM            && \textbf{37.40}{\scriptsize $\pm$0.03} & \textbf{37.33}{\scriptsize $\pm$0.07} &36.36{\scriptsize $\pm$0.05}  & 36.36{\scriptsize $\pm$0.06} \\
\midrule
Mamba + Transf.&SSM + Self-attention          && 37.48{\scriptsize $\pm$0.03} & 37.35{\scriptsize $\pm$0.08} & 36.74{\scriptsize $\pm$0.27} & 36.25{\scriptsize $\pm$0.02}     \\
\textbf{Mamba + Transf.}&\textbf{SSM + LSTM}               && \textbf{37.50}{\scriptsize $\pm$0.07} & \textbf{37.41}{\scriptsize $\pm$0.05} & \textbf{36.94}{\scriptsize $\pm$0.05} & \textbf{36.52}{\scriptsize $\pm$0.08} \\
\bottomrule
\end{tabular}
\label{tab:ablate_neck_anet}
\end{table}

\subsection{Ablation Study on Loss Functions}

In this section, we analyze commonly used loss functions in TAD methods and conduct an ablation study on the effectiveness of the actionness loss proposed for the one-stage method, ActionFormer.

\textbf{Action Category Losses.} These losses are typically classification-based, as action categories are discrete. TAD methods can employ either binary or multi-class classification losses, depending on the objective—distinguishing action from non-action segments or classifying specific action categories. Methods that incorporate external class labels during post-processing (e.g., G-TAD~\cite{xu2020g} and ActionFormer~\cite{zhang2022actionformer} on ActivityNet) typically use binary classification for all category losses within the network. Conversely, methods without such external annotations use multi-class classification for final category predictions while still employing binary classification for intermediate stages.
Common classification losses include focal loss~\cite{Lin2020FocalLF} for binary classification, as used in VSGN~\cite{zhao2021video}, and cross-entropy loss for multi-class classification, as seen in ActionFormer~\cite{zhang2022actionformer}. Some approaches, such as BSN~\cite{lin2018bsn} and BMN~\cite{lin2019bmn}, instead regress action confidence based on the IoU between proposals and ground-truth actions, treating the problem as a regression task rather than a strict binary classification.

\textbf{Action Boundary Losses.} These losses aim to refine the boundaries between predicted and ground-truth action segments, as precise boundary localization is crucial for TAD performance. Various methods improve boundary regression accuracy by directly predicting the distances to start and end locations (e.g., ActionFormer~\cite{zhang2022actionformer}, VSGN~\cite{zhao2021video}) or by regressing offsets relative to predefined anchors, as in PGCN~\cite{liu2020progressive}.

\textbf{Effect of Actionness Loss.} Beyond classification and boundary regression losses, the two-stage method BMN introduces a Temporal Evaluation Module, which classifies each timestep based on actionness/startness/endness to enhance boundary learning. However, this design has been primarily used in two-stage methods. To evaluate its effectiveness in a one-stage setting, we integrate a temporal evaluation head into ActionFormer and apply actionness loss supervision.
The results, shown in Table~\ref{tab:ablateloss}, indicate that while actionness loss provides a minor improvement on ActivityNet, it negatively impacts performance on THUMOS. Given its marginal benefit on ActivityNet and its detrimental effect on THUMOS, we exclude actionness loss from our final design.

\begin{table}[t]
\centering
\caption{\textbf{Analysis of the loss design choices} on ActionFormer.}
\small
\begin{tabular}{ccccc}
\toprule
\textbf{Category} & \textbf{Boundary}   &\textbf{Actionness}  &  \textbf{ActivityNet-v1.3} & \textbf{THUMOS-14} \\
\midrule
\cmark & \cmark     &  & 37.00{\scriptsize $\pm$0.05} & 67.93{\scriptsize $\pm$0.19} \\
\cmark & \cmark     & \cmark & 37.19{\scriptsize $\pm$0.05} & 67.11{\scriptsize $\pm$0.30} \\
\bottomrule
\end{tabular}
\label{tab:ablateloss}
\end{table}

\section{Limitations}

OpenTAD provides a unified framework for implementing and benchmarking various temporal action detection methods, supporting eight datasets. However, it currently focuses exclusively on fully supervised TAD and does not yet support weakly-supervised or open-vocabulary TAD.

Another limitation lies in the scale of existing TAD datasets. Current datasets are relatively small, leading to high variance in experimental results across different random seeds. This variability poses challenges for ensuring training stability and reproducibility. Addressing the scalability of TAD datasets and improving the robustness of training pipelines remains an open research direction that warrants further exploration.

\begin{table}[htbp]
    \centering
    \scriptsize
    \caption{\textbf{Components mapping from each method to OpenTAD.} Cls. and Reg. denote classification loss and regression loss respectively. }
    \rotatebox{-90}{
        \begin{tabular}{llllll}
\toprule
& \textbf{Method} & \textbf{Backbone} & \textbf{Neck} & \textbf{RoI extraction} & \textbf{Losses}       \\
\midrule
\multirow{6}{*}{\begin{sideways}\textbf{One-Stage}\end{sideways}} 
& ActionFormer~\cite{zhang2022actionformer} & I3D, TSP, SlowFast & Multi-scale Transformer &   & Cls. (Focal), Reg. (DIoU)      \\
& TriDet~\cite{shi2023tridet}           & I3D, TSP, R(2+1)D, SlowFast   & Scalable-Granularity Perception  &  & Cls. (IoU weighted Focal) Reg. (DIoU)                  \\
& TemporalMaxer~\cite{tang2023temporalmaxer}  &  I3D, SlowFast  & MaxPool1d &  & Cls. (Focal), Reg. (DIoU) w/ SimOTA                  \\
& VideoMambaSuite~\cite{chen2024video}         & InternVideo2-6B  & Mamba  &  & Cls. (Focal), Reg. (DIoU)     \\
& DyFADet~\cite{yang2024dyfadet}         & I3D, TSP  & Dynamic Feature Aggregation  &  & Cls. (Focal), Reg. (DIoU)     \\
& CausalTAD~\cite{liu2024harnessing}         & I3D, TSP  & Mamba + Transformer  &  & Cls. (Focal), Reg. (DIoU)     \\
\midrule
\multirow{4}{*}{\begin{sideways}\textbf{Two-Stage}\end{sideways}} 
& BMN~\cite{lin2019bmn}     & TSN  & Conv1D  &  Boundary Matching &  Cls. (weighted BLR), Reg. ($L_2$), TEM loss (weighted BCE))  \\
& GTAD~\cite{xu2020g}       &  TSN & GCNeXt & SGAlign   & Cls. (weighted BCE), Reg. ($L_2$), Node Cls. (weighted BCE)) \\
& TSI~\cite{liu2020tsi}     & TSN, I3D, TSP  & Temporal Boundary Detector & IoU Map Regressor & Cls. (Scale-Invariant loss), Reg. ($L_2$), TBD loss (BLR)     \\
& VSGN~\cite{zhao2021video} & TSN, I3D, R(2+1)D & xGPN & Boundary Sampling & Cls. (Focal), Reg. (GIoU), Boundary Adj. (GIoU), Supp. Scores (weighted BLR)\\
\midrule
\multirow{5}{*}{\begin{sideways}\textbf{DETR-Based}\end{sideways}} 
& & & & & \\
& & & & & \\
& TadTR~\cite{liu2022end}   & TSN, I3D & Deformable Transformer & Action Queries, RoIAlign & Cls. (CE), Reg. (IoU, $L_1$), Act. ($L_1$), w/ bipartite matching                \\
& & & & & \\
& & & & & \\
\midrule
\multirow{5}{*}{\begin{sideways}\textbf{End-to-End}\end{sideways}} 
& AFSD~\cite{lin2021learning}       & I3D & FPN & Boundary pooling & Cls. (Focal), Reg. (tIoU, $L_1$)  \\
& E2E-TAD~\cite{liu2022empirical}   & TSN, TSM, I3D, SlowFast &  Same as G-TAD, AFSD, TadTR  \\
& ETAD~\cite{liu2022etad}           & TSM, R(2+1)D &  LSTM & SGAlign &  Cls. (CE), Reg. (smooth-$L_1$)  \\
& Re$^2$TAL~\cite{zhao2023re2tal}   & Re$^2$Vswin, Re$^2$Slowfast & Same as ActionFormer, VSGN      \\
& AdaTAD~\cite{liu2024adatad}       & VideoMAE, SlowFast & Same as ActionFormer   \\
\bottomrule
\end{tabular}
}
\label{tab:mapping_to_opentad}
\end{table}


\end{document}